# Sora and V-JEPA Have Not Learned The Complete Real World Model

-A Philosophical Analysis of Video AIs Through the Theory of Productive Imagination

Jianqiu Zhang
Department of Electrical and Computer Engineering
University of Texas at San Antonio
San Antonio, TX, 78249
Email: michelle.zhang@utsa.edu

**Abstract**

Sora from Open AI has shown exceptional performance, yet it faces scrutiny over whether its technological prowess equates to an authentic comprehension of reality. Critics contend that it lacks a foundational grasp of the world, a deficiency V-JEPA from Meta aims to amend with its joint embedding approach. This debate is vital for steering the future direction of Artificial General Intelligence(AGI). We enrich this debate by developing a theory of productive imagination that generates a coherent world model based on Kantian philosophy. We identify three indispensable components of the coherent world model capable of genuine world understanding: representations of isolated objects, an a priori law of change across space and time, and Kantian categories. Our analysis reveals that Sora is limited because of its oversight of the a priori law of change and Kantian categories, flaws that are not rectifiable through scaling up the training. V-JEPA learns the context-dependent aspect of the a priori law of change. Yet it fails to fully comprehend Kantian categories and incorporate experience, leading us to conclude that neither system currently achieves a comprehensive world understanding. Nevertheless, each system has developed components essential to advancing an integrated AI productive imagination-understanding engine. Finally, we propose an innovative training framework for an AI productive imagination-understanding engine, centered around a joint embedding system designed to transform disordered perceptual input into a structured, coherent world model. Our philosophical analysis pinpoints critical challenges within contemporary video AI technologies and a pathway toward achieving an AI system capable of genuine world understanding, such that it can be applied for reasoning and planning in the future.

## Introduction

OpenAI and Meta, forefront innovators in artificial intelligence, have each introduced their latest video AI technologies, Sora (OpenAI 2024) and V-JEPA (Bardes et al. 2024) , targeting the ambitious goal of achieving "real-world understanding." Such endeavors are not just deemed as technological valuable but also crucial towards the realization of Artificial General Intelligence (AGI), promising AI systems that can engage in sophisticated planning and reasoning by simulating physical realities with precision.

OpenAI's Sora is heralded as a "World Simulator," aiming to replicate complex real world phenomena, and it has garnered a lot of attention for its advanced ability to produce highly realistic videos from textual prompts, an achievement some researchers argue signifies a leap towards AGI (Trabert 2024). It utilizes text annotations, refined by a state-of-the-art captioning model using a technique similar to Dall.E's re-captioning approach (Betker et al. 2023). Through diffusion models, Sora learns the statistical distributions of videos, enabling video generation conditioned on specific text prompts. Nonetheless, OpenAI notes certain limitations: Sora's simulation of complex physical interactions and its grasp on cause-and-effect remain underdeveloped (OpenAI 2024).

Prominent figures in AI research, such as Yan LeCun, have voiced skepticism about equating current generative AI capabilities with AGI. LeCun argues that true understanding of the world extends beyond generating compelling video samples from text prompts (LeCun, 2024; LeCun, 2022). He advocates for a focus on "world models" based methods aiming to model the latent variables essential for a genuine understanding of the world . World model based video AIs utilize techniques including variational auto-encoders (VAEs), transformers, and joint embedding predictive architectures (JEPA) (Ha and Schmidhuber 2018; Robine et al. 2023; Chen et al. 2022; Hafner et al. 2019; Bardes et al. 2024). Meta's newly released V-JEPA (Bardes et al. 2024) is designed to learn the correlation of video patches across space and time, which is deemed as a form of “true” understanding of the world.

However, the real-world applicability of such “world model” based video AIs remains in exploratory stages. V-JEPA, for instance, has been trained using short video segments from a massive dataset of 2M video clips. Its performance has been accessed on tasks like action recognition and motion classification. Despite achieving satisfactory results on simpler datasets such as Kinetics400 (K400) (Kay et al. 2017) and Something-Something-v2 (SSv2) (Goyal et al. 2017), V-JEPA's performance is still poor when applied to the AVA dataset (Gu et al. 2017), which presents complex scenarios involving multiple actors and actions.

The rapid advancements in video AI technology, exemplified by OpenAI's Sora and Meta's V-JEPA, have brought the concept of "real-world understanding" into sharp focus. Discussions revolve around key questions: Can Sora's expanded training datasets address its problems and achieve genuine world understanding? Is V-JEPA's methodology a game-changing breakthrough? Could a combination of these technologies provide a comprehensive framework for modeling real-world phenomena? Establishing criteria for what constitutes a genuine "real" world model—identifying the critical elements it must include—is central to this discourse.

To tackle these complex discussions, philosophical inquiry is essential. This inquiry should clarify the nature of the data inputs and outputs, and the overarching goals of video AI technologies. Key questions include:

- **Characteristics of Perceptual Data:** How do the characteristics of video clips used in AI training align with human perceptions of the world? What are the distinct features of these clips?
- **Sources of Input:** What types of input are crucial for building a world model? How do these sources influence the model's development?
- **Purpose and Functionality:** What are the objectives of creating a world model from this data? Is the primary function of the model to recognize objects, or does it focus on identifying abstract, non-representational variables?
- **Representation Requirements:** What types of representations are necessary to support the functionalities designated for the world model?
- **Grounding to Reality:** How can the world model be validated or checked against actual reality to ensure its accuracy and relevance?
- **Comparison with Human Cognition:** Which aspects of human cognitive abilities are mirrored in the processes of video AI models? How do contemporary video AIs stand in comparison to human cognitive structures and functionalities in terms of design and effectiveness? By comparing video AIs to human cognitive processes, we can uncover both the capabilities and limitations of these technologies.

These questions go beyond the usual scope of technology-focused research, which typically concentrates on a limited set of specific objectives. For example, Sora is mainly designed to train a statistical model using diffusion-transformers to create video clips from textual prompts or other inputs. Yet, the broader context of perception and understanding encompasses various factors and objectives that shape the development of a world model. Thus, a philosophical examination becomes essential to address these issues comprehensively, which not only delves into the technical functionalities of AI but also explores how these technologies mirror complex human cognitive processes.

Given these considerations, the philosophical framework outlined by Immanuel Kant in his "Critique of Pure Reason"(Kant 1908) becomes relevant for our investigation into the nature of video AIs in the larger context of human perception and understanding. Kant proposes that our perceptions are triggered by external stimuli that activate our sensory organs. These perceptions are inherently fragmented: we do not perceive the entire world simultaneously but rather begin with specific parts and explore others in a seemingly random time sequence. From this, Kant deduces the necessity of an internal mental process that synthesizes these scattered and disjointed inputs into a coherent, unified perception, which he calls "the unity of the manifold of intuition," facilitated by what he terms the "productive imagination" (CPR, A123). This mental function, according to Kant, is crucial not only for the basic perception of objects but also for imparting meaning and coherence to our sensory experiences, thereby foundational to object recognition (CPR, A146).

Applying Kant’s insights to video AIs provides clear answers to some of our questions. First, Kant’s observation that our perceptions are fragmented is also true for video clips: they offer a disorganized view of the world from human perspectives, albeit with greater editorial freedom. In videos, scenes from different perspectives can be stitched together, but they must still adhere to logical rules. We can recognize illogical sequences when they contradict these established rules. Second, Kant's concept of "the unity of all manifolds of intuition" can be likened to what we describe as a "coherent world model" in AI systems. Drawing from this comparison, we see that the role of a coherent world model extends beyond merely sequencing time-series data. It also involves enabling AI to accurately identify objects and their relationships, which is crucial for attaining a realistic comprehension of the world. The development of such coherent world models in AI is intricately linked to the process of understanding, which entails the integration and interpretation of fragmented sensory data in a manner that reflects human cognitive functions.

Kant's exploration of productive imagination introduces four crucial categories of pure understanding recognized through schemata—innate, rule-based frameworks that precede empirical experience and shape our comprehension of time, sequences of events, and the overarching concept of time as it relates to experiential phenomena (CPR, B185). For instance, within the relation category, Kant illustrates that the schema governing cause and effect posits a rule-based sequence where specific conditions reliably lead to certain outcomes (CPR, B184). This exemplifies the critical role of category in temporal event comprehension.

Kant's categories are crucial for object recognition. This is clearly illustrated in the optical illusion where one might see either two faces or a vase, depending on the perceptual interpretation. The Kantian category of "quantity" plays a decisive role in this cognitive process—if perceived as a single object, it is seen as a vase; if seen as two, then as faces. This example highlights the significant role that Kantian interpretative frameworks play in guiding how we recognize objects.

While Kant delineates these categories as fundamental a priori conditions for cognition, he stops short of claiming an exhaustive list, thereby acknowledging the potential for additional a priori conditions necessary for a nuanced understanding of time and event sequences. This opens up a rich area for philosophical debate and interpretation of Kant's theory.

Predictive Processing (PP) is another relevant theory in cognitive science which posits that perception is constructed by the brain by testing hypotheses about the hidden causes of incoming sensory stimulation, and the hypotheses are generated by a hierarchical generative model of the world which is constantly updated based prediction errors (Rao and Ballard 1999; Friston 2003, 2005, 2008, 2010; J. Hohwy 2010; Clark 2015). Some researchers argue that this theory has conceptual ground with Kantian philosophy (Swanson 2016), particularly in its interpretation of perception forming—an active process characterized by the synthesis of intuition and concepts, as opposed to a mere building up of percepts from sensory inputs (Gładziejewski 2016).

PP encompasses a range of paradigms, from unsupervised learning methods that build models of sensory inputs by learning from previously learned image patterns (Rao and Ballard 1999) to modeling sensory data as an auto-regressive process (Friston 2008). Additionally, there are supervised learning approaches that connect complex symbols to sensory data within a generative framework (Santana and Principe 2015; Principe and Chalasani 2014). Note that world model based Video AIs typically align with the unsupervised paradigm of PP, including systems that adhere to unsupervised temporal predictive coding frameworks (Ha and Schmidhuber 2018; van den Oord, Li, and Vinyals 2018) and those that utilize unsupervised spatial predictive coding or a combination of both spatial and temporal approaches (Doersch, Gupta, and Efros 2015; Zhang, Isola, and Efros 2016; LeCun and Courant 2022; Bardes et al. 2024).

The challenges faced by PP include, firstly, its failure to recognize the problem of fragmented perceptual data. Secondly, the lack of a unified PP framework that integrates its various paradigms highlights the theory's conceptual fragmentation. Moreover, ongoing debates concerning the philosophical foundations of PP, especially its stance on representationalism, complicate its application as a comprehensive theoretical framework for analyzing video AIs, as discussed in works by (Gładziejewski 2016; Facchin 2021).

Therefore, while PP presents an intriguing intersection with Kantian theory and offers additional insights to understanding video AIs, its practical application in analyzing video AI technologies like Sora and V-JEPA requires more precise conceptual development given the significant unresolved issues within the PP framework, and it becomes crucial to analyze these video AI technologies using the broader concepts of a coherent world model and productive imagination that we propose.

In pursuing this analysis, we choose not to delve deeply into debates of Kantian philosophy or PP theory interpretation, but rather to use their foundational concepts as a springboard for proposing a modified theory of productive imagination. For historical context and a review of previous applications of Kantian philosophy in AI, readers can refer to (Schlicht 2022). To assist those unfamiliar with Kantian terminology, we clearly defined critical concepts such as “unity of manifold of intuitions,” “productive imagination,” or “schemata”.

As we revisit the initial questions about video AIs, it becomes clear that while Kant’s theory provides answers to some, many remain unaddressed: He elucidated the fragmented and chaotic nature of our perceptual inputs. He also identified the function of productive imagination in forming a coherent world model (the unity of manifold of intuitions), preparing it for object recognition through applying the schemata of Kantian categories. However, he left other questions unanswered. To leverage Kant's insights for the analysis of video AI technology, it is crucial to enhance his theoretical model to encompass all the important elements outlined in our list of questions.

In this paper, we developed a novel theory of productive imagination, applying a Kantian analytical method often referred to as "reverse engineering" or "top-down" analysis(Swanson 2016). This method critically assesses the input and output of AI systems to deduce the functionalities of their intermediate processing stages.

Our analysis indicates that productive imagination synthesizes fragmented perceptual data with experience to develop coherent world models. This position contrasts with Kantian theory, which overlooks the role of experience in the creation of coherent models. Additionally, this perspective differs from the PP theory, which posits that perception is entirely constructed from pre-existing models.

Our research indicates that productive imagination can handle dual input sources and function in various modes, serving diverse purposes. It can track reality, react to unrecognizable objects, simulate scenarios or generate illusions in a dreaming mode, and blend past experiences with current perceptual data to enhance understanding. In comparison, neither Kant's theory nor PP explicitly addresses the use of multiple input sources, the variety of operating modes, or the multifaceted purposes of our visual perception and understanding engine.

Essential to this process is the correct temporal organization of objects and events within fragmented perceptual data, and the application of Kantian categories. We introduce an a priori law of change, reflecting Kant's concept of transcendental affinity (CPR, A113), which governs the changes of objects' movements, shapes, and appearances over space and time. By establishing spatial and temporal contexts and clarifying object relationships, the coherent world model represents objects and their associated events in time in a context, enabling the cognition module to recognize them and engage associated experience, which allows further refining the model.

Echoing PP, our theory includes a reality check module, a feature not present in Kant's framework. This module uses the coherent world model to predict future scenes and then compares these predictions to actual perceptual data. Any discrepancies, likely stemming from biases in experience, lead to adjustments in the model, ensuring that it remains closely aligned with reality.

Our analysis addresses all aspects of the inquiring questions and leads to a complete analytical framework for describing the productive imagination-understanding engine. We then extend our philosophical framework to analyze the two prominent video AI systems, Sora and V-JEPA, revealing that both systems, despite their innovative approaches, fail to address key aspects of productive imagination. Sora, operating in a "dreaming mode," uses a diffusion-transformer as an experience composer, generating videos from text prompts from past experiences or creating illusions based on the prompt's realism. However, its failure to integrate Kantian categories and the a priori law of change limits its ability to organize perceptual data correctly in time, supporting LeCun's critique that generating convincing videos does not equate to genuine understanding. V-JEPA, while capturing some aspects of the a priori law of change by learning video patch correlations, also overlooks Kantian categories and lacks the capability to incorporate experience, which limits its field of application and its performance in complex scenarios.

In summary, while both Sora and V-JEPA lay foundational groundwork for an AI perceptual-understanding framework, neither is adequate for achieving AGI. For improvements, both systems should incorporate Kantian categories and the a priori law of change to enhance object recognition and relation understanding, integrate all three operational modes of productive imagination, and employ mechanisms to organize and utilize experience effectively, ensuring a coherent and temporally accurate world model.

To address the challenges, we propose a new architecture designed to fully develop an AI productive imagination-understanding engine. This architecture involves training a coherent world model encoder based on fragmented, out-of-order video clips. The goal is to ensure that the output from this encoder matches the output of an encoder trained on non-fragmented, correctly ordered video clips.

This paper makes contributions to the field of video AI in three key areas: 1. It provides a philosophical analysis and outlines a detailed methodology for transforming fragmented perceptual data into coherent world models and cognitive frameworks. 2.
It examines two leading video AI systems, demonstrating that although each captures certain elements effectively, neither succeeds in achieving a truly coherent world model; 3. It introduces a Coherent World Model Learning Architecture (CWMLA) which showcases how the training of coherent world models can be deeply rooted in philosophical principles, particularly those espoused by Kant.

Our analysis leads to a deeper, philosophically-grounded understanding of existing video AI systems and facilitates the creation of advanced AI systems genuinely equipped for real-world comprehension, essential for effective planning and reasoning based on accurate world simulations. The findings from this research could significantly influence future AI development strategies, guiding substantial investments towards achieving breakthroughs in AGI.

## Section 1: A Theory of Productive Imagination

### Section 1.1 Coherent world model

Kant characterized productive imagination as a crucial intermediary faculty that bridges sensory perception and intellectual understanding.  One primary function of productive imagination is to synthesize fragmented and disordered perceptual data into a coherent world model. Additionally, it is also tasked with isolating individual objects from their contexts, tracking their changes over time, and establishing their interrelationships, which are essential before they can be recognized by understanding. This process allows contingent perceptual data to be associated with specific defined concepts, forming the basis for subsequent AI actions.

Kant identified four essential categories of pure understanding necessary for constructing this coherent world model, split into two groups: the mathematical categories, which include the quantity and quality of objects, and the dynamical categories, which cover the objects' inter relationships and modalities. These categories need to be universally applied across all experiences to facilitate object recognition.

We argue that the Kantian categories function to separate objects from their surroundings, determine their properties, and recognize their relations and interactions with their environment for the purpose of ordering and identify the events and objects contained in the perceptual data: Segmenting perception into distinct objects allows us to evaluate their quantity (Unity, Plurality, Totality). By understanding the properties of objects, we can assess their quality ( Reality, Negation, and Limitation). For example, hardness is real in solid objects, fluids have no shape, and transparency is limited by purity of materials. Understanding object modality (Possibility, Existence, Necessity) constraints the status of object existence across time under changing context. For example, it is necessary that a force is present if an object is moved. It is possible that a solid object will break into pieces if dropped, and there must exist a container if liquid doesn't flow. Finally, recognizing an object's relationships to others (Inherence and Subsistence, Causality and Dependence, Community) helps determine how dynamically changing objects are related, and establish the timeline of their interactions. For example bumping one solid object into another causes the second to move, a predator and a prey's interaction form a community. These categories collectively address the question: What are the distinct objects, their generic properties, and how can they be isolated in space and time for recognition? What might happen given the context and their own properties? How are they related to each other in time events?

According to some Kantian scholars, the categories are innate in the input/output sense: "even though their acquisition may have been occasioned by sensory stimuli, its content does not derive from sensory stimuli, but it is contributed by the mind (Vanzo 2018)." Given this interpretation, video AIs may acquire categories through training and abstraction of existing patterns in data.

Kant explains that categories of pure understanding are recognized through schemata: "The schemata are therefore nothing but a priori time-determinations in accordance with rules, and these concern, according to the order of the categories, the time-series, the content of time, the order of time, and finally the sum total of time in regard to all possible objects. (CPR, B185)." This ordering of events over time is critical for a coherent experience and is necessary for object recognition within the understanding process.

The determination of isolatable objects and their temporal arrangement are two intertwined goals. Without identifying discrete objects (content of time), it's impossible to establish a sequence of events (order of time); conversely, without recognizing the sequence, the content cannot be accurately determined. This suggests that understanding both aspects requires an iterative process where time content and time order are mutually informative.

The exploration of Kantian categories sheds light on key components required for a coherent world model that enables understanding. Notably, the proper sequencing of time is fundamental and impacts all other cognitive functions. However, the task of organizing fragmented, disordered data into a logical sequence remains a substantial challenge. In the next section, we introduce and develop the concept of an a priori law of change, crucial for establishing the correct temporal order.

**Section 1.1.1 The Deduction of the a priori law of change**

Kant posits that during the synthesis of perceptions, there exists 'a rule by which a representation is conjoined in the imagination with one representation rather than with another' (CPR, A121). He explained that “the representation of a universal condition according to which a certain manifold can be uniformly posited is called a rule, and, when it must be so posited, a law. Thus all appearances stand in thoroughgoing connection according to necessary laws, and therefore in a transcendental affinity, of which the empirical is a mere consequence.“ (CPR, A114). However, the concept of "transcendental affinity" is not comprehensively explained, and it is unclear exactly what this law entails and how an AI system could potentially learn it. The primary focus of this paper is to explore the specific contents of this law rather than delve into detailed interpretations of Kantian theories. For a thorough examination of Kant's views on this matter, please consult the detailed analyses available in the referenced works (Westphal 2005; Griffith 2012; McLear 2015).

Kant's insights indicate that the primary role of productive imagination is to construct a timeline of objects and events from fragmented perceptual data. Transcendental affinity dictates how this timeline should be structured, yet the specifics of this rule remain unclear. To better understand this rule, we can reconsider the problem from a different angle: once a timeline is formulated from disjointed perceptions, the isolation and contextualization of objects within this timeline facilitate the modeling of their visually detectable changes in movement, appearance, and shape. Essentially, in creating a timeline and isolating objects within it, models of object changes over space and time are simultaneously established. Consequently, the rule that governs the formation of timelines and time-content determination is the same rule that regulates these models of object change, evaluating the plausibility of timelines and setting constraints on the acceptable changes of objects over time. Since this rule is the basis for synthesizing fragmented perceptual data, it embodies transcendental affinity, and we refer to it as the **a priori law of change** over space and time, highlighting its role in governing object changes in movement, appearance, and shapes.

The next crucial question concerns the relationship between this law and Kantian categories, as both serve as a priori conditions for the creation of a coherent world model. Are they independent or interconnected? Does the a priori law of change override the Kantian categories? Our analysis suggests that while the a priori law of change complements Kantian categories, it remains distinct from them. On one hand, the properties of objects influence how the a priori law of change is applied; on the other, this law itself helps define certain Kantian categories. This interplay indicates that while they are interrelated, each maintains its unique influence within the framework of constructing a coherent world model.

For example, a rigid object retains its shape, whereas a pliable one may deform, thus necessitating different applications of the a priori law based on the material properties, a factor within the Kantian category of quality. Conversely, Kantian categories can sometimes be derived from this law, as shown in the infant number-tracking experiment conducted by Xu and Carey in (1996). In this study, a column and a ball behind a screen were pushed out and shown one by one before returning behind the screen. The a priori law of change dictates that the column cannot suddenly transform into a ball behind the screen, leading infants in the experiment to infer the presence of two distinct objects. This illustrates how the a priori law can guide the understanding of the Kantian category of quantity.

The a priori law of change is also context-dependent, reflecting Kant's category of modality, where the possibilities and necessities of events are shaped by the interactions between objects and their surroundings. For instance, a solid object will drop in mid-air is a necessary outcome within that context. Conversely, the shattering of a glass when it hits a hard surface represents a contextual possibility. Yet, invariant object properties like color do not influence this law of change, indicating the clear distinction between the categories and the law.

**Section 1.1.2 Coherent world model and experience**

Kant did not explicitly address how experience is integrated into the "unity of manifold of intuition." Nevertheless, this integration is readily apparent: observing the front of a house naturally leads us to infer its backside. Our perceptions and accumulated experiences converge to form a unified mental model, an integration that hinges on our experience and current perceptual data. The extent of this integration varies with our familiarity with the objects in a scene. At one extreme, lacking prior experience, productive imagination must rely solely on perceptual input to construct a coherent world model. At the other extreme, even in the absence of perceptual data, we can envisage a world drawn from our well of experience.

Consequently, we conclude that productive imagination draws from two input sources: fragmented perceptual input and experience. With different configuration of inputs, it can operate in three distinct modes: 1. Reality tracking mode, in situations devoid of experience, 2. Dreaming mode, where imagination is based solely on past experiences without new sensory input, and 3. Mixing mode, where both perceptual data and experience inform the construction of the world model.

None of the three modes of productive imagination is purely passive that only respond to incoming sensory information; each requires the active participation of the mind. For example, even in reality tracking mode, the mind actively engages the a priori law of change and Kantian categories to synthesize information.

Regarding the structure of experience, it can be categorized into three types: reproductive, compositional, or a combination of both. Reproductive experience is derived directly from memory. In contrast, compositional experience is formed from conscious or unconscious thoughts. The nature of compositional experience can range from realistic to illusionary, depending on the rationality of the thoughts that initiate the composition process. In the context of artificial intelligence, this compositional process is akin to the function of the diffusion-transformer in Sora which produces video clips based on “thoughts” in the form of text prompts. In the following discussion, we refer to all types of experience without differentiating the subtypes.

In summary, the coherent world model encompasses latent variables of at least three key aspects: those associated with isolated objects, those detailing object changes across space and time constrained by the a priori law of change and experience, and those pertaining to Kantian categories. In future sections, we will evaluate the completeness of various video AI world models based on these aspects.

**Section 1.2. Anchoring of productive imagination to reality**

In the preceding discussion, we've delineated that productive imagination must synthesize fragmented perceptual data and experience to construct a coherent world model. However, to ensure that such a model does not diverge into fantasy all the time, there must be a process that grounds it in reality.

The validation of the coherent world model against reality can only occur through comparison with new, incoming perceptual data because the current model is derived from previous perceptual data and experience. Consequently, for the reality check, the current model must align its prediction with fresh perceptual input. When there's a strong correspondence, the model can be considered accurate.

This process entails a balance between experience and perceptual data: The outcome of this reality check influences the weight placed on experience. A discrepancy would necessitate a greater reliance on current perceptual data, minimizing potential biases inherent in experience.

Our understanding of how productive imagination is grounded to reality resonates with the PP theory, which asserts that ‘‘The sensory input to the brain does not shape perception directly: sensory input is better and more perplexingly characterized as feedback to the queries issued by the brain’’ (Jakob Hohwy 2013) . In other words, the current perceptual data is used as a comparison to the model that PP constructs such that it can adjust its model to fit the reality. While both PP and our perspective endorse the use of world model predictions for reality checking, our theory uniquely focuses on calibrating the balance between perceptual data (formed directly from sensory stimuli) and experience, whereas PP aims to refine the model to better fit the sensory inputs to form perception.

Given different degrees of adherence to reality that is permissible in our theory of productive imagination, it's crucial to distinguish between the predictive and the dreaming mode. Prediction is specifically aimed at reality testing, whereas dreaming serves no such purpose. Predictions are made when productive imagination integrates current perceptual data into the coherent world model, necessitating a immediate-future focus for reality checks. Dreaming, in contrast, leans heavily on experience when perceptual data is absent. An extended forecast into the future resembles a dream more than an immediate prediction. Between these extremes lies a spectrum where prediction and dreaming intermingle.

### Section 1.3: The productive imagination-understanding engine

In the previous two sections, specific aspects of productive imagination are addressed. This section takes a holistic view of the entire productive imagination-understanding engine. Kant suggested that the process from sensory stimuli to understanding progresses linearly, starting with sensory intuition (perception), progressing through imagination (for forming coherent world models), and culminating in apperception (for recognizing objects). However, for the coherent world model's predictions to be validated against new perceptual data—a process defined as the reality check—recognition of objects must already be in place. Without this recognition, it is not possible to utilize experience about identified objects or correct cognitive mistakes. Therefore, an iterative loop that integrates productive imagination with understanding is essential: Initially, fragmented perceptual data and/or experience are used to create a coherent world model. This model facilitates object recognition, which then allows the integration of associated experience and new perceptual data to refine the model during the reality check. This cycle of model updating and reality checking is repeated with each new iteration.

The a priori law of change offers a foundational framework of physical laws, abstracted from specific object details, which enables the productive imagination to transform disordered perceptual input into a structured, coherent world model. Knowledge of specific objects, gained through experience, subsequently enriches this model after recognition has taken place. This integration of general physical laws with experience introduces both objective and subjective elements into the world model, with the latter prone to biases and inaccuracies. Consequently, a reality check is essential to keep the productive imagination anchored in reality. Without such verification, the cognitive process risks drifting into ungrounded, dream-like illusions.

An overall illustration of the productive imagination-understanding engine is shown in Figure 1.

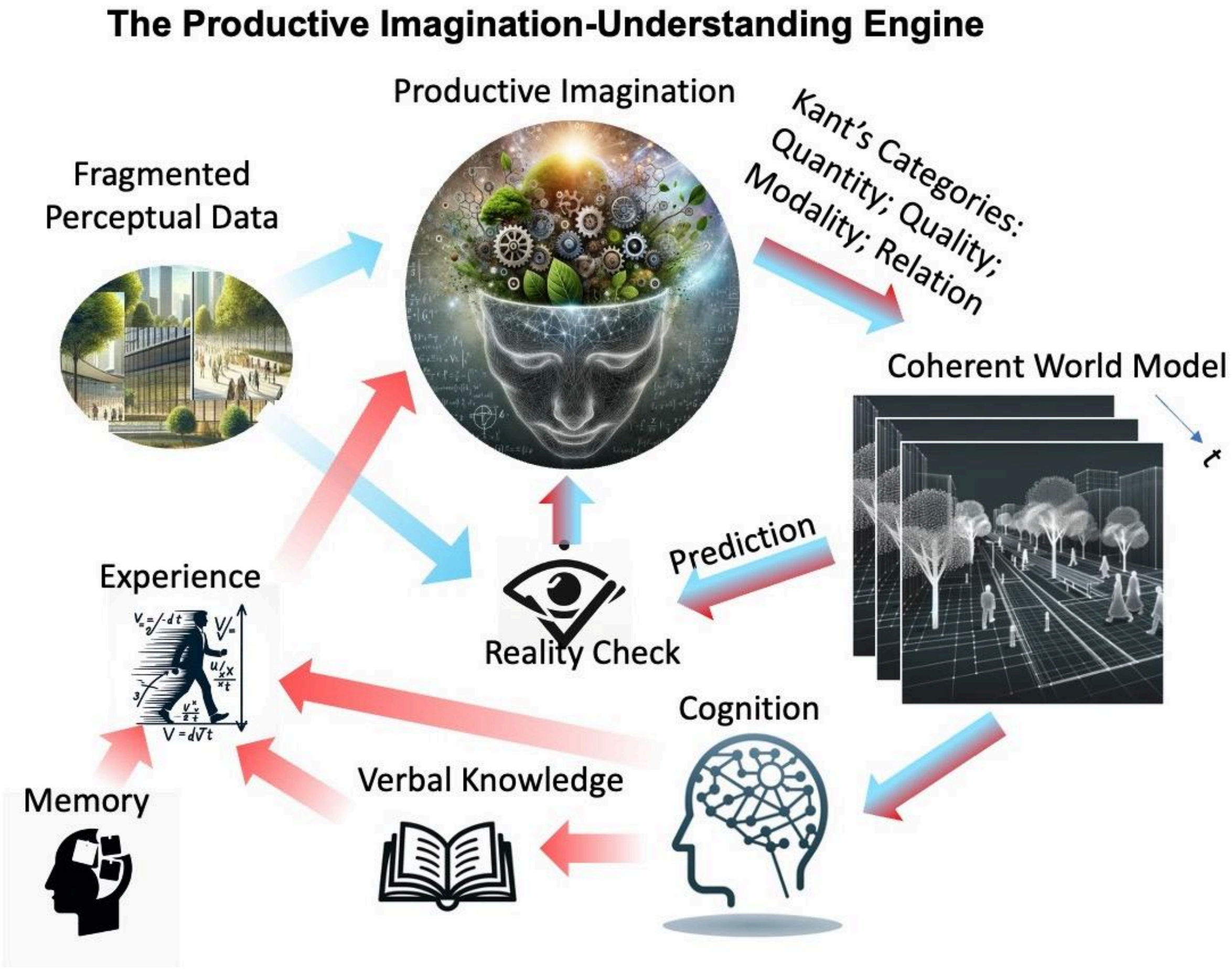

Figure 1. The Central Role of Productive Imagination in Understanding. Blue arrows represent the flow of objective reality based on perceptual data. Red arrows depict the flow of subjective knowledge derived from object recognition and experience. Arrows combining red and blue signify the information flow from a blend of objective and subjective sources.

In Figure 1, the inputs for productive imagination—perceptual data and experience—are represented by blue and red flows respectively. Each can independently activate productive imagination, while a combination of both is depicted with blue and red arrows. This flexibility allows for operation in various modes:

**Reality Tracking Mode**: When driven solely by perceptual data, productive imagination aligns its coherent world model with the incoming perceptual data through an "objective" feedback loop. This loop comprises productive imagination, the coherent world model, future predictor, and reality checks. This mode is dominant when object recognition is minimal, suppressing the influence of associated experience. The a priori law of change and Kantian categories can still be applied to reach an understanding of the scene, allowing activities such as object tracking and path following.

**Dreaming Mode**: In the absence of perceptual data, productive imagination draws exclusively from experience to compose scenes, akin to dreaming. This mode supports real world simulation if one thinks realistically or relies on memory. However, because the engine in the dreaming mode is unfettered by reality checks, fantasy and creativity becomes possible if thinking is unconstrained by logic or memory, including Kantian categories and the a priori law of change.

**Mixing Mode**: Situated between reality tracking and dreaming, the mixing mode synthesizes both perceptual data and experience, which is activated by the recognized objects and their associative links to other objects, derived from verbal knowledge or direct experience in memory.

In the mixing mode, the interplay of perceptual input, experience, Kantian categories, and the a priori laws of change impose constraint on the possible changes of objects. In contrast, the reality tracking mode allows for novel experiences, while the dreaming mode can transcend constraints like common logic, sidelining even the a priori laws and Kantian categories to enable fantasy.

In summary, the process of understanding involves productive imagination merging fragmented perceptual data with experience to construct a coherent world model. This model produces predictions that are validated through a reality check, leading to updates that enhance further recognition and understanding. This continuous cycle facilitates the understanding of objects and their interactions within environments. The result is a detailed and coherent world model that distinguishes objects and elucidates their relationships in their context, drawing on both perceptual data and experience.

**Section 2: A diagnosis analysis of Sora's world model**

Recently released by OpenAI, Sora represents a significant advancement in the field of video generative AI technologies. Detailed historical perspectives on the evolution of video generative AIs can be found in the studies by (Liu et al. 2024; Cho et al. 2024). In this paper, we select Sora as a focal example of the latest advancements in generative AI, highlighting its exceptional performance that potentially represents a significant step toward developing an operational world model. This capability is crucial for allowing AI to simulate realistic scenarios and plan on future actions, as suggested by Sora's description as a "world simulator" in OpenAI's technical documentation (OpenAI 2024).

This analysis does not intend to delve into the intricate technical details of Sora's implementation as discussed by Liu et al. (Liu et al. 2024), instead, we offer a high-level overview to assess its structure and identify any limitations. Sora leverages diffusion models and is trained on video clips that are annotated with text. These annotations are produced by a sophisticated captioning model that is trained using a re-captioning technique similar to that employed in Dall.E (Ramesh et al. 2022), which generates Descriptive Synthetic Captions (DSC), providing detailed descriptions that include not only the main subjects but also background elements and detailed attributes of each object.

OpenAI's technical documentation provides an overview of the Sora framework, which includes a front-end video compression network that condenses video clips into latent variables (Kingma and Welling 2013; Rombach et al. 2022). These latent variables are then broken down into space-time patches, which are utilized as input tokens for transformers. Sora employs a novel diffusion transformer approach (Peebles and Xie 2023), replacing the conventional U-Net architecture with transformers to effectively capture the distribution of input images conditioned on text captions.

A high-level diagram of Sora's training framework is drawn in Figure 2, styled similarly to Figure 1, to provide clarity on how the system compares to the human productive imagination-understanding engine.

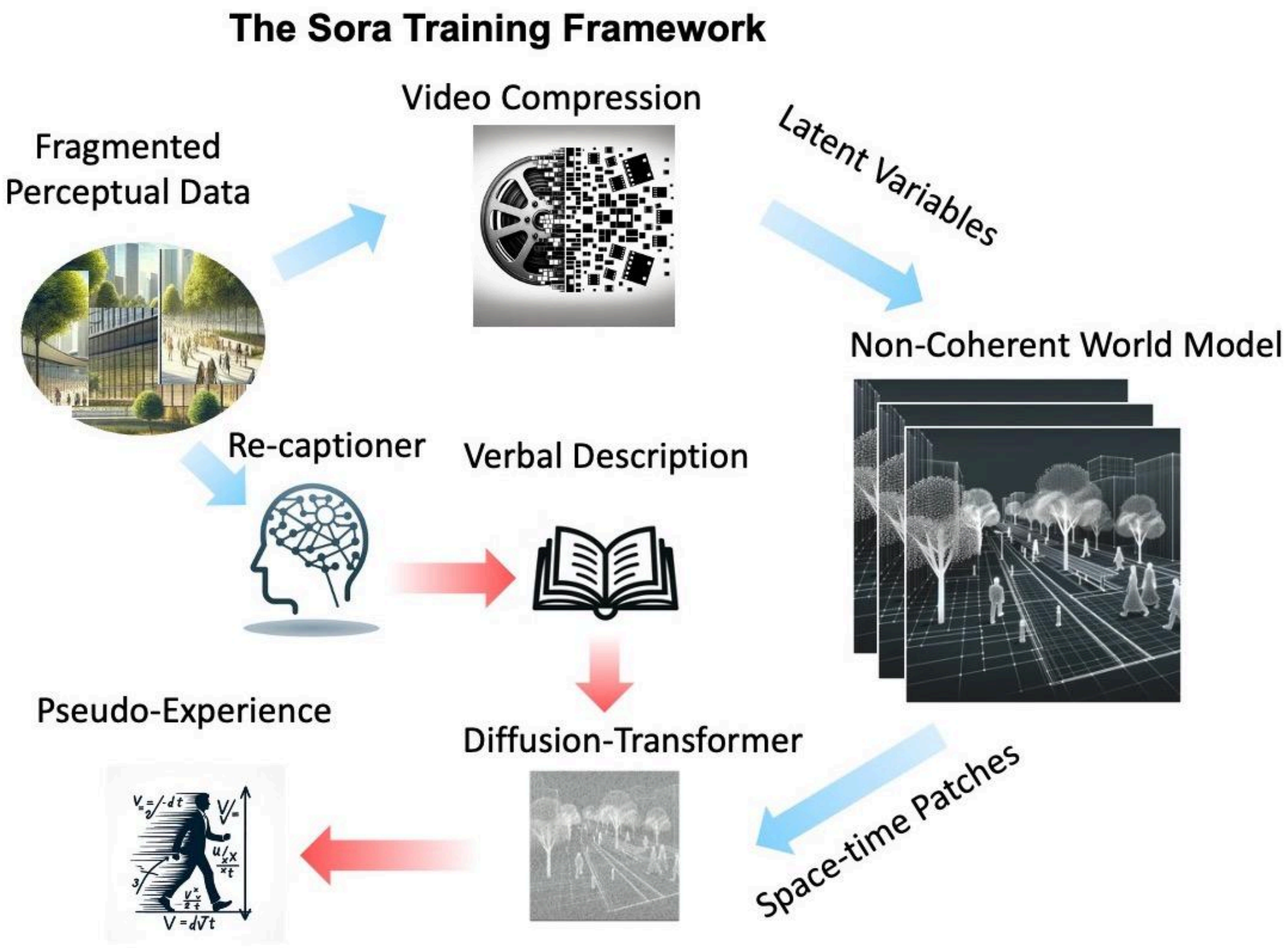

Figure 2: The current training framework of sora

In Figure 2, the training process starts with fragmented perceptual data from video clips, which undergoes video compression, transforming the videos into latent variables, which provide a compressed representation that captures certain elements of the world model. It is important to note that this process does not incorporate any a priori laws of change or Kantian categories of understanding. Consequently, the model generated through this compression can only be described as a "Non-Coherent World Model."

These latent variables, which represent the non-coherent world model, are then converted into space-time patches, which are processed by a diffusion-transformer, a deep neural network designed to capture the probability distributions of the patches based on the text descriptions accompanying them. If we compare the function of the diffusion-transformer  with components that interact with productive imagination in Figure 1, it can be elucidated that the diffusion-transformer is equivalent to a repository of experience in function. When prompted with text, it composes and generates space-time patch samples from the video clips' conditional distribution that align with the text description.

To train a genuine coherent world model, the input to the diffusion-transformer should include all outputs of productive imagination and cognition including latent variables representing coherent world models such as the Kantian categories and the associated experience of the recognized objects. However, comparing Figures 1 and 2 in terms of the list of elements, it is evident that while some Kantian categories may have been incorporated into the DSCs, there is a lack of systematic incorporation of all Kantian categories. Other coherent world model latent variables and genuine experience are also absent from the training process.

Upon completion of training, the Sora system is capable of generating 60-second video clips that appear realistic sometimes. This generation process is detailed in Figure 3.

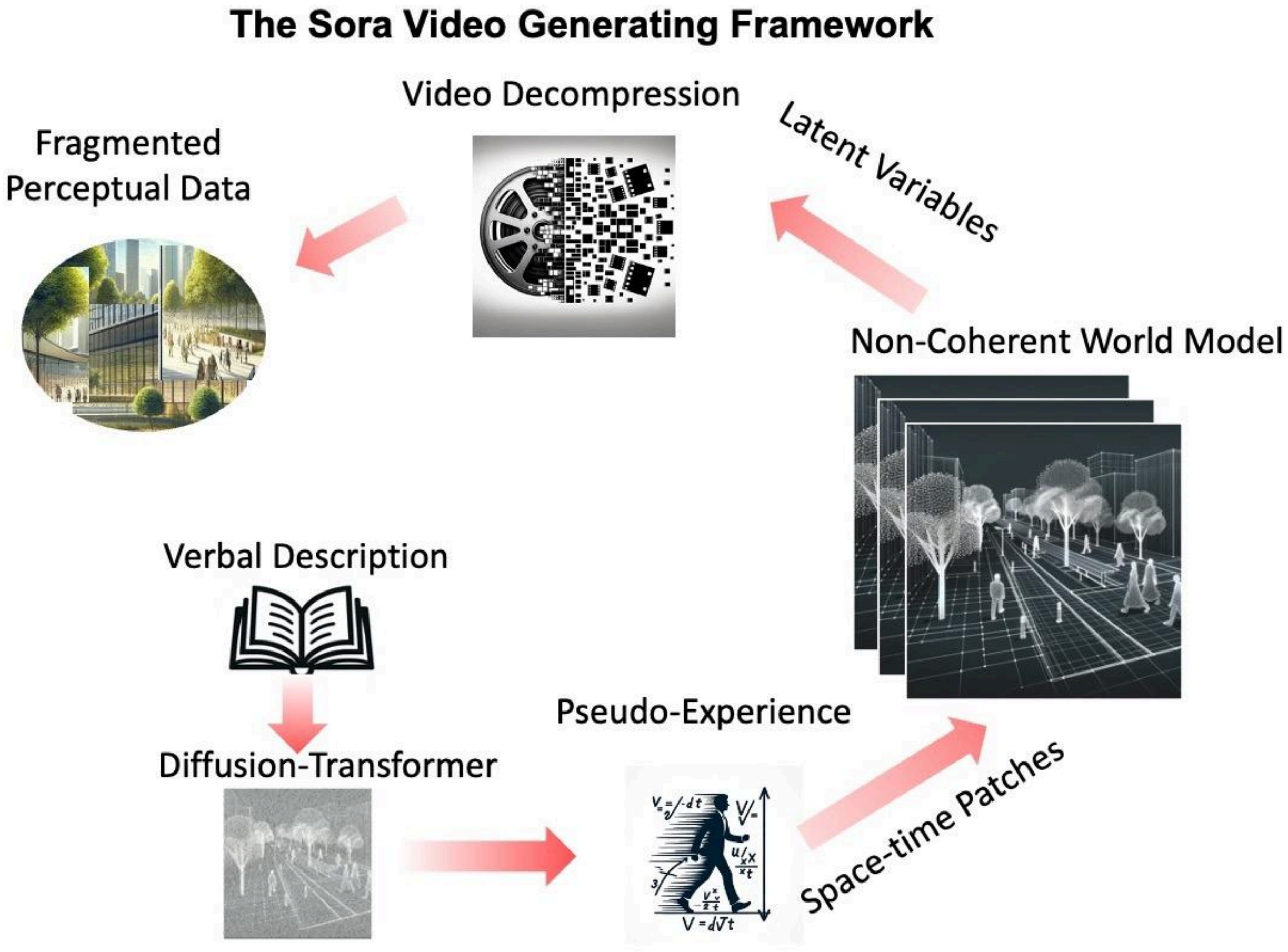

Figure 3: The video generation process of Sora. All arrows are red to indicate that all information flow comes from subjective sources in a dreaming mode.

The video generation process in Sora begins with text prompts fed into the diffusion-transformer. From these prompts, a sample of video clips as space-time patches is produced, reflecting a non-coherent world model. These patches are then converted into latent variables, which, after being processed through a decoder/decompressor, result in a video clip corresponding to the initial text prompt.

Sora demonstrated composing capability in creating non-realistic objects and scenes from text prompts. It implies that Sora has learned distinct latent variables representing individual objects, allowing it to creatively combine elements, such as generating a hybrid creature from a giraffe and a flamingo (donalleniii 2024). Sora's composability suggests that latent parameters related to individual objects have been captured, based on which non-existent creatures become possible.

However, Sora-generated videos also exhibit significant flaws. By comparing the video generation process illustrated in Figure 3 with the productive imagination-understanding engine in Figure 1, it's evident that Sora operates similarly to the dreaming mode of productive imagination—relying solely on text prompts and associated experience to compose scenes, without real-world perceptual data to anchor its outputs in reality.

Since Sora does not construct a coherent world model during its training, the experience it gathers is deficient in temporal ordering. Additionally, free composition in the dreaming mode often results in deviations from physical laws due to a lack of constraint from the a priori law of change and Kantian categories . These shortcomings manifest as several identifiable flaws in the videos it generates:

**Fragmented Perception**: The generated videos assemble fragmented perceptual data not unified in a coherent world model.
**Inconsistency in Kantian Categories**: Since Kantian categories were not integrated into the Descriptive Synthetic Captions (DSCs) or any other aspect of training, Sora struggles to consistently maintain these categories in its scenes. For example, without understanding the category of quantity, an arbitrary number of objects may appear. Without a grasp of the category of causality, the Sora fails to generate a logically ordered series of events from cause to effect.
**Violation of Physical Laws**: The a priori law of change is not learned during training, leading to illogical changes in object movement, appearance, and shape over space and time.
**Lack of Common Sense**: The lack of genuine experience based on correct time ordering can result in scenes that do not adhere to common sense.

These deficiencies align with observations from Sora's output. For instance, in a video from the Sora technical report (OpenAI 2024), a glass is depicted as lifting off from a desk, changing shape, and spilling its contents before touching down again, which showcases errors in physical representation, sequence timing, and causality. Such errors indicate that the training process did not incorporate necessary constraints on object change over time, nor did it correctly sequence events, reflecting a fundamental misunderstanding of basic physical and causal relationships. Some of the defects are shown in Figure 4.

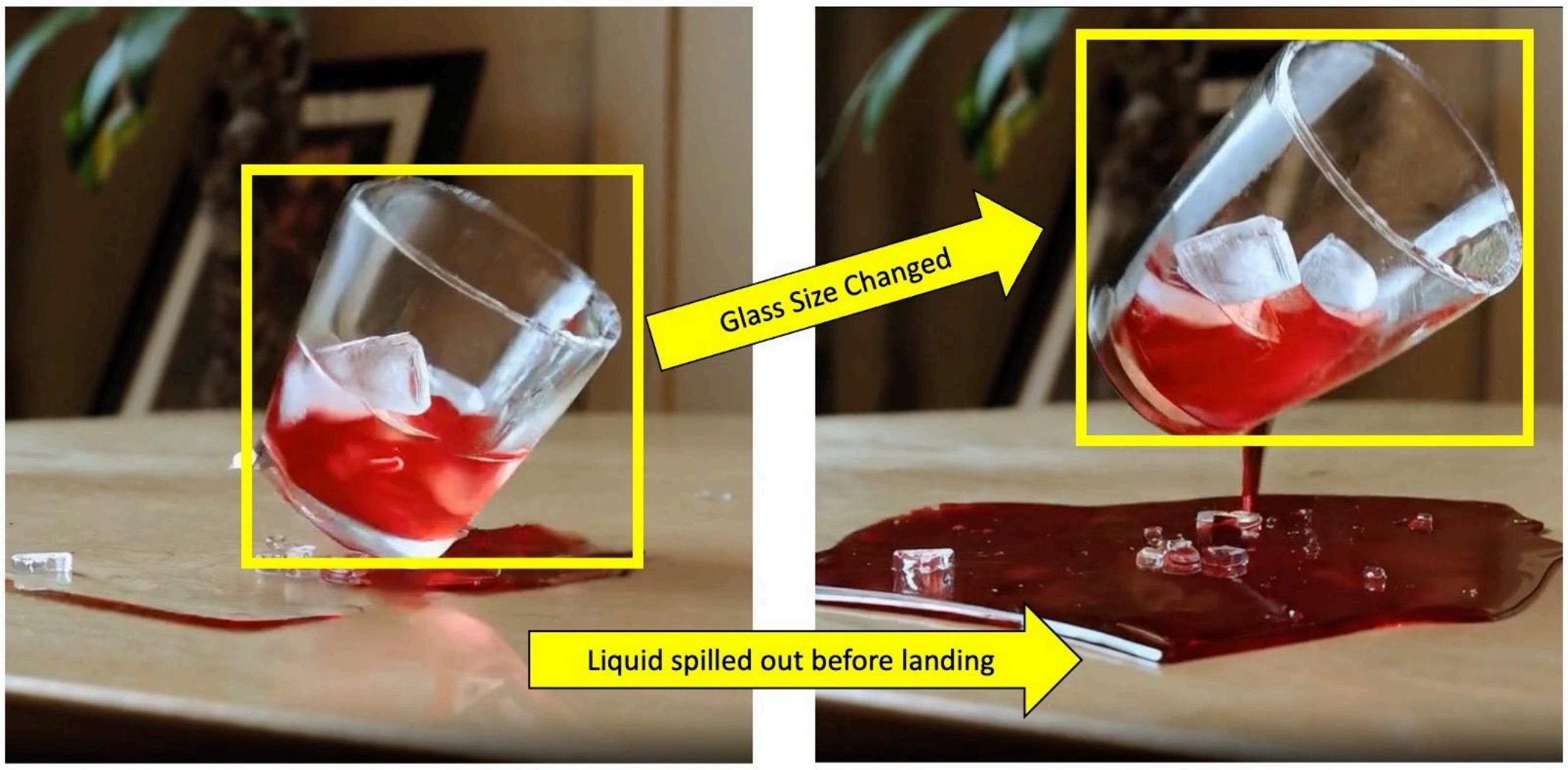

Figure 4: Defects in the glass shattering video: 1. Glass growing bigger indicates that the a priori law of change has not imposed the correct constraint on shape change 2. Liquid spilling out before landing indicates that the category of causality was not applied properly when forming the world model. 3. The upward movement does not follow physical laws, indicating a lack of constraint from genuine experience.

On the Sora introduction page (https://openai.com/sora), five videos are showcased, each illustrating different issues. One video features a man walking on a treadmill that inexplicably drifts from side to side, suggesting a failure to implement the a priori law of change, which should anchor the treadmill in one spot once it's decided that the treadmill is a heavy object that shouldn't move. As a result, the representation resembles fragmented perceptual data.

In another video, intended to depict five gray wolves playing, the scene begins with three wolves and inexplicably shifts to include four or more, without any visual cue of new wolves entering the scene. This abrupt change fails to maintain the Kantian category of quantity derived based on the a priori law of change, which shall exclude the possibility of one wolf suddenly becoming two.

A third video involves a basketball passing through a hoop and similarly demonstrates objects randomly appearing, further indicating a disregard for the Kantian category of quantity. Additionally, a second ball passes through the rim, revealing a lack of a priori law of change that shall prohibit solid objects passing through other solid objects.

The fourth video depicts archaeologists discovering a plastic chair that randomly changes shape. This error suggests a misunderstanding of the chair as a solid and non-flexible object and a failure to apply the a priori law of change that should govern the stability of solid objects' shapes.

Lastly, a video shows an elderly woman blowing out candles on a birthday cake, but the flame remains unaffected by her action. This clearly demonstrates a misunderstanding of the category of cause and effect, as the expected outcome of extinguishing the flames after blowing them does not occur.

Some of these problems are illustrated in Figures 5 and 6, highlighting the deficiencies in Sora's video generation process concerning fundamental physical and logical principles. There are more examples of similar problems that have been documented in (Cho et al. 2024) .

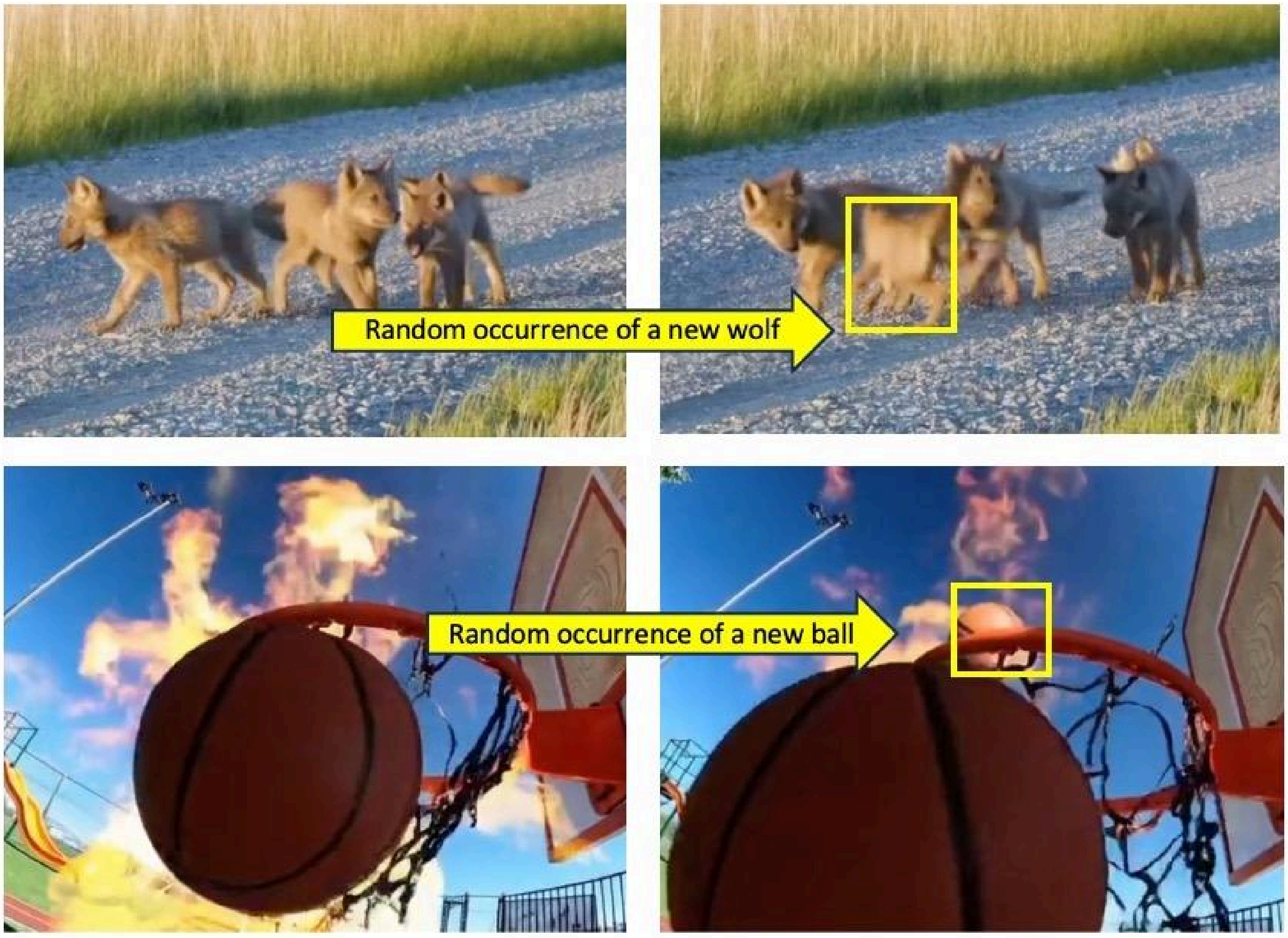

Figure 5: Random occurrence of new objects into the scene indicates that the category of quantity is not applied when setting up the world model.

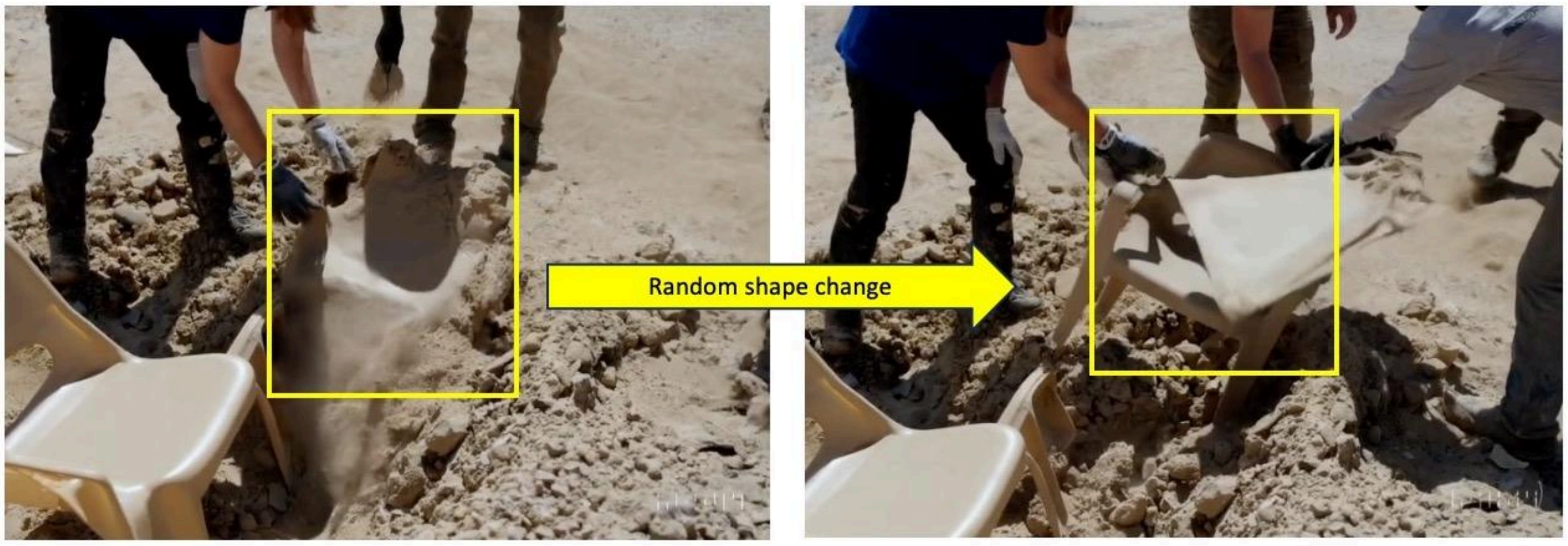

Figure 6: Random change of the plastic chair's shape indicates that the a priori law of change is not applied, and the chair is not understood as an inflexible object.

In summary, Sora generates a non-coherent world model based on pseudo-experience that lacks appropriate time-ordering and consequently the understanding of physical laws during its training process. Sora operates predominantly in what can be described as the dreaming mode, using text prompts without any grounding in reality. On the positive side, Sora captures latent variables of isolated objects, enabling the composition of novel entities and scenes within this mode. The shortcomings of Sora stem from the absence of an AI equivalent of a coherent world model. This model should ideally include latent variables related to the a priori law of change and the Kantian categories of pure understanding.

Without these critical constraining variables, Sora is unable to adhere to the physical laws governing object movement and transformation, leading to the various issues observed in its outputs. These problems are not just technical glitches that can be resolved by scaling up training; they signify a fundamental flaw in the Sora architecture. To move towards a realistic world simulator—a significant step toward achieving AGI—a profound architectural overhaul of Sora is required.

**Section 3. A diagnostic analysis of V-JEPA**

In February 2024, Meta introduced V-JEPA (Video-Joint Embedding Predictive Architecture) that takes a distinct approach toward constructing realistic AI world models. Unlike more immediately applicable AI innovations, V-JEPA did not garner significant media attention. This model, outlined in (Bardes et al. 2024) is based on the JEPA methodology conceived by (LeCun and Courant 2022). The core principle of JEPA involves predicting missing patches in partially masked video clips within an abstract latent space, a concept related to contrastive predictive coding as described by (van den Oord, Li, and Vinyals 2018).

V-JEPA's architecture consists of three interconnected networks: a context encoder, a target encoder, and a predictor. The target encoder processes entire video clips, while the context encoder works with masked versions of these clips. Following the context encoder, the predictor generates tokens for the missing video patches. The networks are trained to minimize discrepancies between the predicted and actual missing video patches in the latent space, thus allowing the target encoder to effectively model the world. Each encoder is built using Vision Transformers (ViT) (Dosovitskiy et al. 2020) , employing standard transformer blocks with joint spacetime attention, capable of processing and outputting d-dimensional embedding vectors for each token.

In Figure 7, we illustrate the V-JEPA architecture using a style that is similar to that in Figure 1.

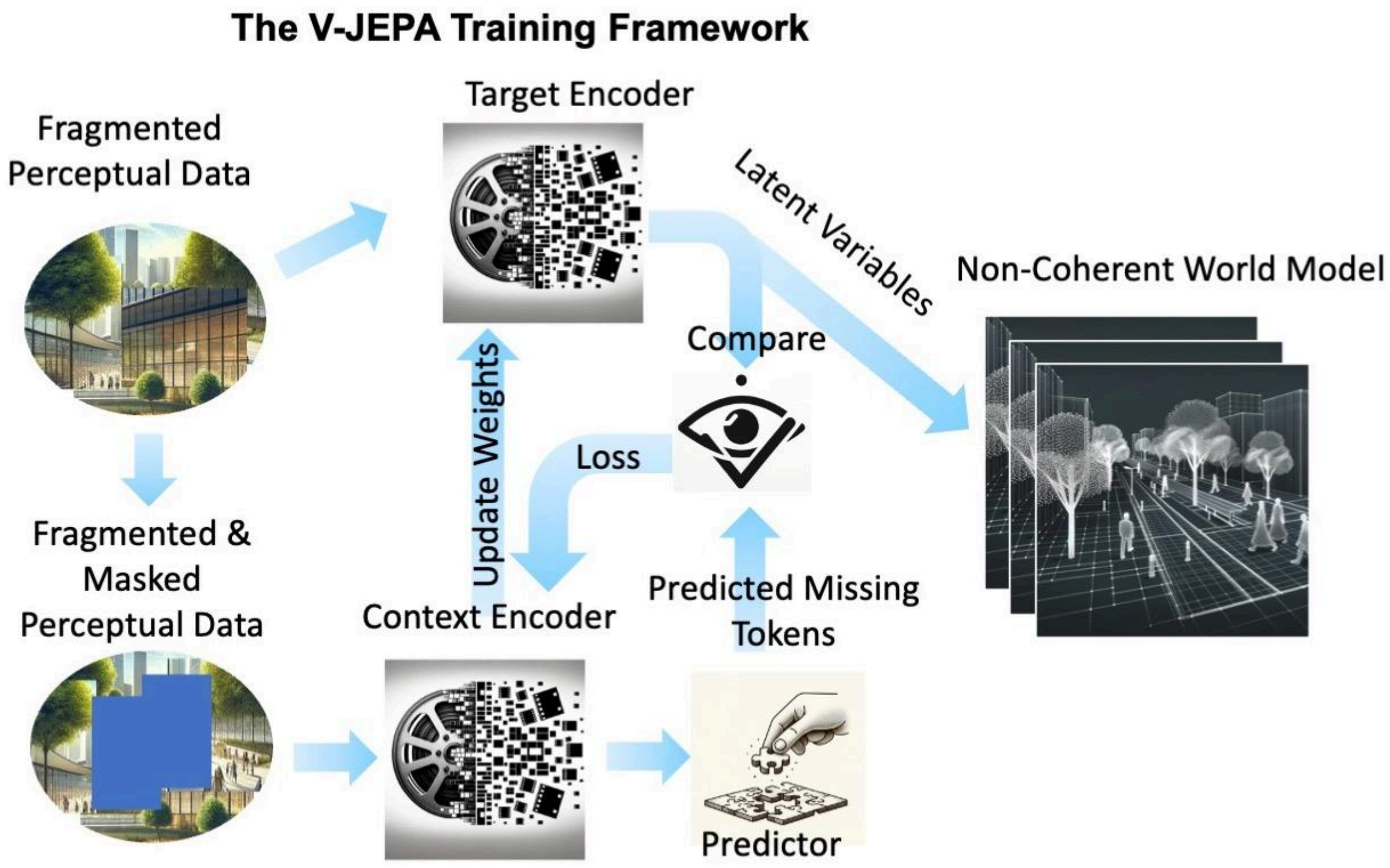

Figure 7: The V-JEPA Training framework.

V-JEPA is trained on a dataset of 2 million publicly sourced videos, each sampled to 64 frames, with the duration of training clips approximately 2 seconds each. To address spatial and temporal redundancy, a 3D Multi-Block Masking technique is applied, effectively masking about 90% of the video content. Consequently, the context encoder processes only around 10% of the video data.

The training methodology involves comparing the outputs from the predictor to those of the target encoder using the sum of the L1 norm across all masked tokens as the loss function. This loss is then utilized to update the weights of the context encoder and the predictor. The weights of the target encoder are adjusted based on the Exponential Moving Average (EMA) of the context encoder's weights.

In the application phase, as depicted in Figure 8, the target encoder is fixed and used to encode incoming videos. The resulting latent variables are input into lightweight probing networks designed for downstream tasks such as classification. This includes an attentive probing protocol (Chen et al., 2022; Yuan et al., 2023) , featuring a simple cross-attention module with a learnable query token and a linear classifier. The effectiveness of this setup is assessed across several public datasets, focusing on the accuracy of object identification, action recognition, motion classification, and action localization.

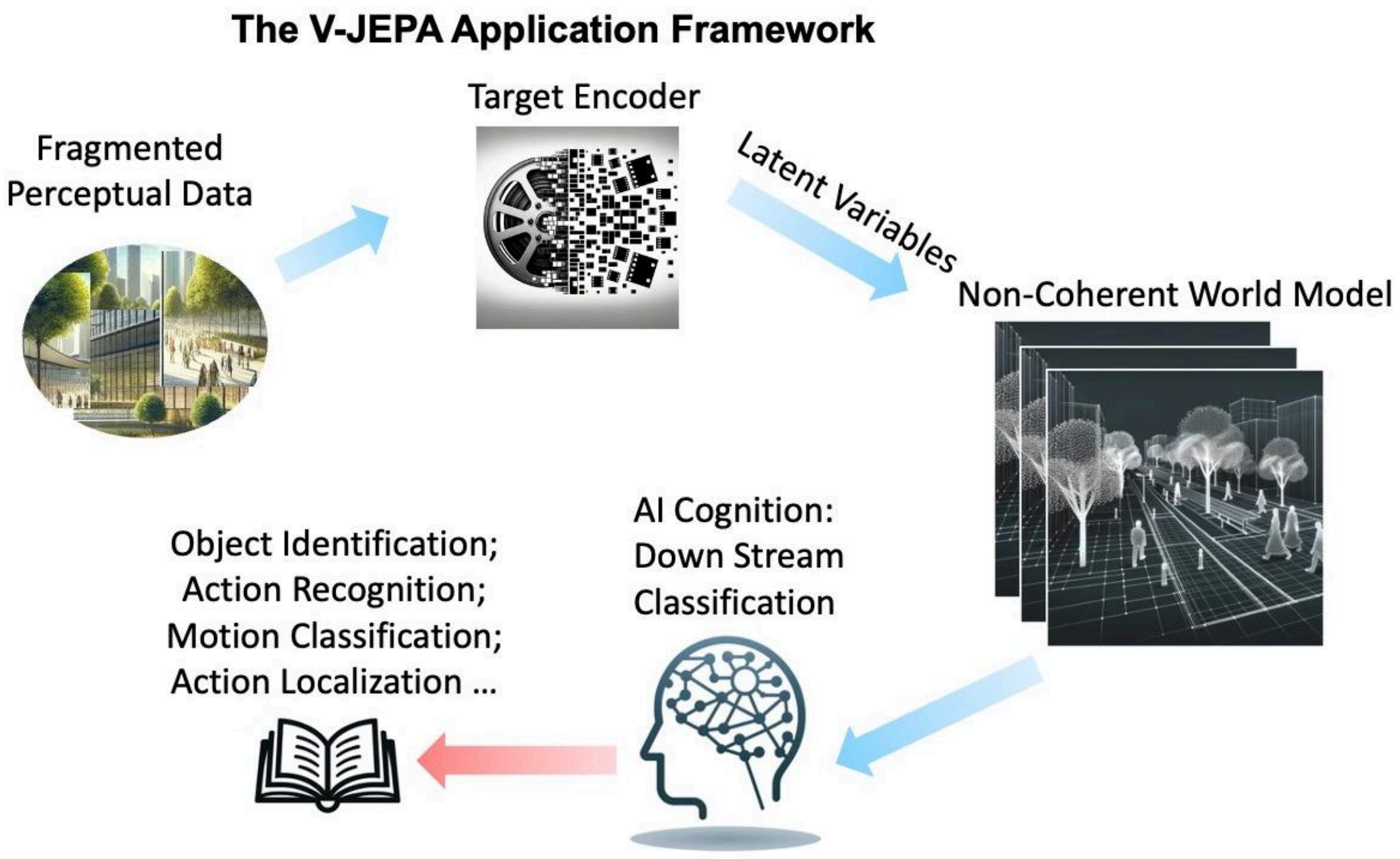

Figure 8: The V-JEPA application framework.

We evaluate the world model learned by V-JEPA from three critical aspects that define a coherent world model. Firstly, by predicting missing patches in videos, the target encoder learns spatial-temporal correlations between unmasked and masked patches. This prediction capability stems from the contextual information and object properties evident in unmasked patches, reflecting the context dependent aspects of a priori laws of change and also the Kantian category of modality. However, due to the brief 2-second duration of the training clips, the model may not capture longer-term correlative features.

Secondly, the training of V-JEPA does not explicitly incorporate all Kantian categories, limiting its capability to fully understand multiple objects and their interactions over time. This limitation is evident in its performance across various datasets: while V-JEPA achieves an 82% accuracy in action recognition on the K400 dataset and 71% on the SSv2 dataset, both featuring relatively simple scenarios,  its accuracy drops to 25% on the AVA dataset, which requires recognizing multiple actors and actions without contextual clues. This indicates V-JEPA's difficulty in distinguishing multiple elements in space-time, and the short video duration further restricts its ability to recognize relational dynamics such as cause and effect.

Thirdly, the model's learning is confined to correlational relationships without evidence of understanding individual object properties or demonstrating composability as seen in Sora. Additionally, V-JEPA lacks the ability to simulate or use experience, which limits its potential.

In conclusion, although V-JEPA captures some spatial-temporal correlations, it does not fully capture the coherent world model. It manages to implement a fraction of the productive imagination needed to transform fragmented perceptual data into a comprehensive world model. Missing elements include reality checks, object recognition, and the use of experience, making V-JEPA a non-generative video AI in contrast to models like Sora operates in the dreaming mode. Thus, while Soar and V-JEPA each contribute to AI understanding, they only implement one aspect necessary for a complete AI understanding system.

## Section 4. A proposed framework for AI Understanding based on coherent world models

In the preceding sections, we develope a theoretical framework explaining how productive imagination could synthesize experience and fragmented perceptual data to forge a coherent world model which enables recognition of objects and their relations to each other in the appropriate context . Central to this framework is a reality check module, which acts as a calibration tool to ensure the model accurately tracks reality. Should discrepancies arise, the model is adjusted by giving greater emphasis to real perceptual data.

Contrasting our sophisticated theoretical mechanism, current state-of-the-art video AI systems fall short in achieving a coherent world model and integrating genuine experience. For instance, Sora struggles with consistently recognizing causal relationships in the correct time sequence, understanding physical possibilities for change in movement, appearance, and shape over time, and maintaining consistency in the number of objects within scenes. Similarly, V-JEPA exhibits subpar performance in action localization within the AVA dataset, particularly in scenarios involving multiple individuals engaged in diverse actions.

To remedy these shortcomings, it is necessary to construct and maintain an equivalent to the coherent world model during training, with text captions that encapsulate all relevant descriptions of Kantian categories. This model is foundational for humans to experience the world within one consciousness, and once established,  it will facilitate the storage and subsequent utilization of authentic experiences to refine the coherent world model in a continuous flow.

We propose a learning architecture for initially training a coherent world model generator, drawing on elements from both V-JEPA and Sora. The core concept involves aligning the outputs from a video encoder processing fragmented and out-of-sequence clips with those from an encoder handling non-fragmented, correctly ordered clips of the same scenes. This proposed architecture is detailed in Figure 8.

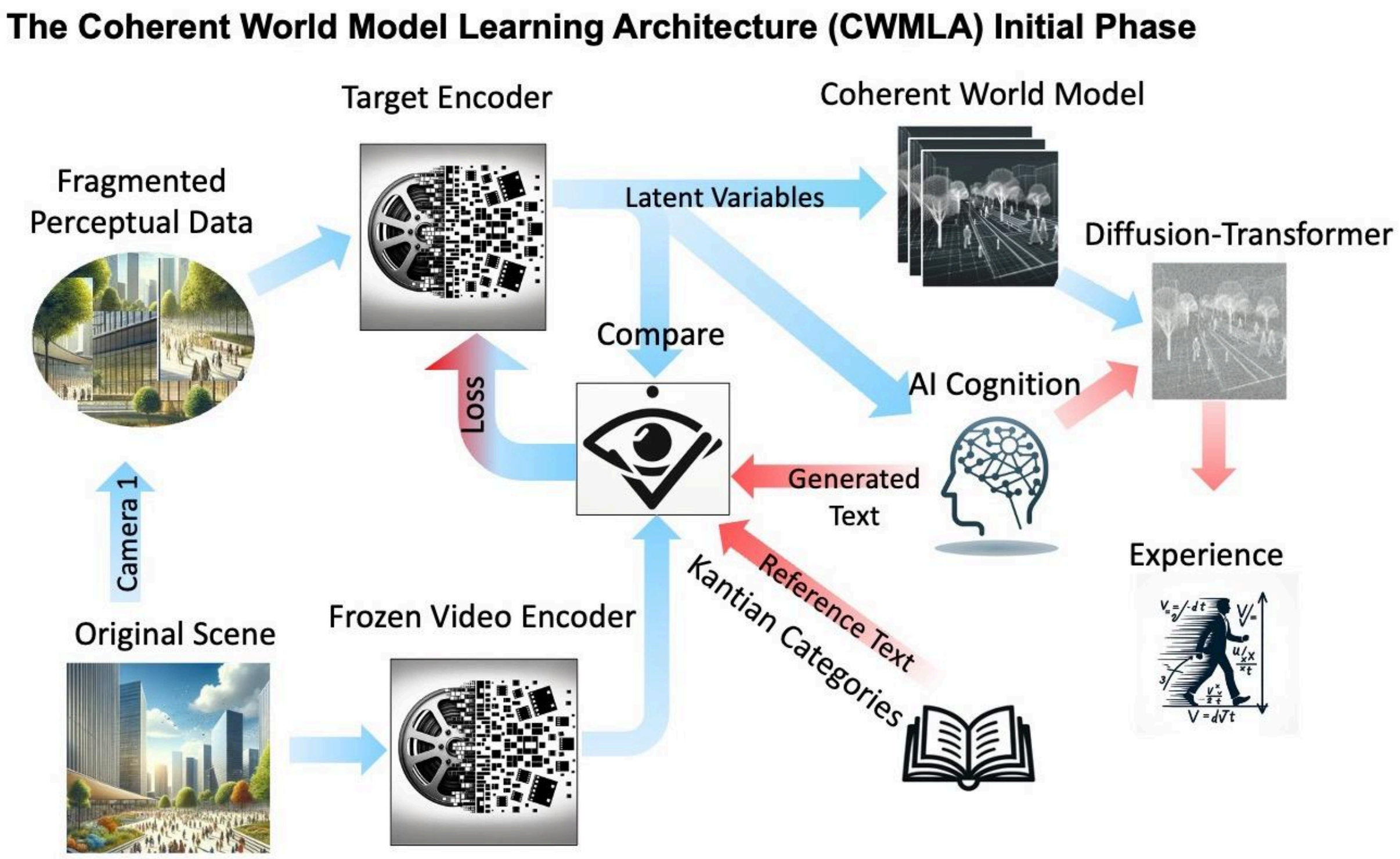

Figure. 8 The proposed initial training architecture for Coherent World Model

The learning process aims at training a target encoder that turns fragmented perceptual data into latent variables representing a coherent world model. The outputs from this target encoder are compared to those from a pre-trained encoder that processes the videos of the original, orderly, and complete scenes. The frozen video encoder, potentially adapted from Sora's video compressor, shall capture latent variables representing individual objects effectively. By aligning outputs from the disordered input with those from the ordered inputs, the target encoder is compelled to learn how to synthesize disordered and fragmented perceptual data into a coherent world model.

To enhance the capability of the target encoder, an AI cognition module is trained alongside it. This module leverages video descriptive captions as training materials, encompassing detailed descriptions of all objects in the scene and relevant Kantian categories articulated through text. Such training ensures accurate representation of both the content and the temporal sequence of events. Incorporating text-based training is vital because the a priori law of change should be co-determined with Kantian categories such as quality and modality. For instance, the constraints on shape changes vary with objects' flexibility. Text descriptions detailing flexibility help train the encoder to recognize object quality and the associated constraints on object changes simultaneously.

It's important to recognize that the a priori law of change may not fully resolve all time-ordering issues through visual cues alone. In real-life situations, humans often rely on additional sensory inputs like heat or smell to understand sequences of events—such as turning on a thermostat followed by removing jackets. In these scenarios, video AI systems may find it challenging to determine the sequence of events relying only on the a priori law of change. This limitation can be mitigated if the AI systems are able to integrate verbal experience that accurately reflects the correct sequence of events in their world model.

In theory, grasping the correct sequence of events and object recognition automatically implies the learning of all elements in the coherent world model. However, in practice, if the target encoder struggles to learn all elements at once, it may be more feasible to train separate encoders for different elements of the coherent world model. The outputs of these encoders can be merged for enhanced processing. Our framework presents an architecture that comprehensively meets the needs for creating coherent world models, though the specifics of implementation will require further refinement by future AI engineers.

Once the initial training phase concludes, the weights of the target encoder can be fixed, enabling the generation of coherent world models as latent variables. These variables, combined with reference texts, shall be fed into a diffusion-transformer, capturing the statistical variations and thus encoding genuine experience.

Following the initial training phases, we obtain a target encoder that effectively converts fragmented perceptual inputs into coherent world models, properly ordered in time. The encoder's outputs embed Kantian categories and parameters related to the temporal dynamics of objects that enforces the a priori law of change.

Following the initial training phase, the final training phase aims to construct a comprehensive AI system mirroring the human productive imagination-understanding engine. The architecture for this phase is depicted in Figure 9.

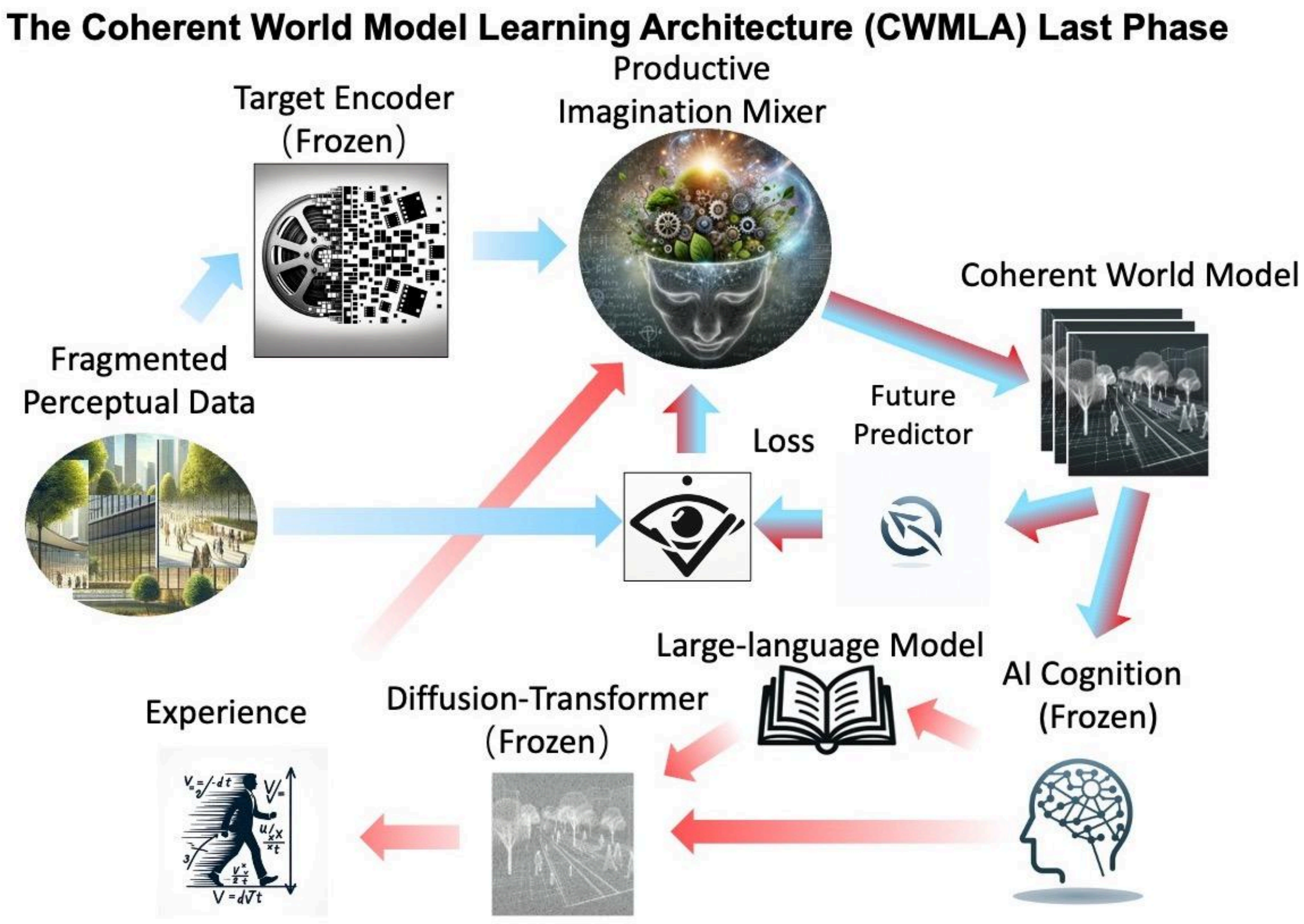

Figure 9 The last phase of Coherent World Model learning

In Figure 9, the target encoder trained in the initial phase will transform fragmented perceptual data into latent variables representing coherent world model parameters. These outputs are subsequently fed into the trained AI cognition unit that outputs textual description of the coherent world model. This unit's outputs, in turn, can be integrated with Large Language Models (LLMs) to access related verbal experience. The combined data, including recognized objects and generated text after accessing LLM, can be fed to the diffusion-transformer trained earlier. The output from this transformer represents experience, effectively sampling latent variables under input constraints. Then experience, together with the latent variables from the target encoder, can be utilized to train a productive imagination mixer that synthesizes both input sources to construct a coherent world model.

The productive imagination mixer is trained through a loss function comparing predicted scenes against actual perceptual data, using a future predictor trained to convert coherent world model parameters into forthcoming perceptual data.

The iterative nature of the training process allows for continuous refinement. Although the initial phase lacks a reality check, post the final training phase, the diffusion-transformer can be updated to better reflect reality and integrate experience.

After several iterations, all units can be consolidated to function in the three modes typical of human productive imagination. The operational framework is the same as illustrated in Figure 9, except that all AI units are frozen..

It is crucial that the diffusion-transformer be continuously updated based on the evolving coherent world models and associated textual references to maintain and refresh experience.

This holistic AI framework, which integrates perception, imagination, and understanding, holds vast potential for diverse applications. In its dreaming mode, it can operate as a world simulator, similar to Sora but augmented with coherent world model representations, which supports AI in planning, reasoning, and creative endeavors. In modes such as reality tracking or mixing, the system assimilates new experiential data, boosting its learning capabilities. Additionally, the interaction between the cognition unit and Large Language Models (LLMs) enables the AI to handle progressively more complex reasoning tasks, utilizing extensive verbal knowledge to tackle upcoming challenges.

**Conclusion**

In this paper, we first underscore the significance of Kant's theory of productive imagination for analyzing modern video AIs like Sora and V-JEPA, particularly in their capacity to develop coherent world models. We emphasize the importance of Kantian categories for isolating objects, determining their properties, modalities, and relationships prior to achieving understanding. However, we also point out the gaps in Kant's theory, including the absence of an iterative process for ongoing correction of cognitive errors and reality tracking, explicit mechanisms for enforcing accurate temporal sequencing, and the integration of experience. To remedy these deficiencies, we show that the ***a priori law of change*** *across space and time* is critical for establishing temporal order. We discuss how Kantian categories and the a priori law of change are interconnected yet distinct. Moreover, we describe an architectural framework for a productive imagination-understanding engine that includes both an objective reality tracking loop and a subjective dreaming loop. We elucidated three operational modes of productive imagination—reality tracking, dreaming, and mixing—and identify three essential types of latent variables: those related to individual objects, those reflecting changes in objects governed by the a priori law of change over space and time, and those associated with Kantian categories.

Applying this theoretical framework, we analyze Sora, noting its operation akin to the dreaming mode of productive imagination. The analysis is able to show that Sora's defects stem from its inability to sequence fragmented perceptual data correctly, incorporate Kantian categories for proper object relation understanding, and enforce the a priori law of change. Our analysis also covers V-JEPA, a video AI system developed by Meta, recognizing its ability in capturing contextual aspects of the a priori law of change. However, its performance is limited in complex action localization tasks because it does not integrate Kantian categories nor experience into its world model. We conclude that while these two leading video AIs do not fully realize the coherent world model, they contribute elements vital to a comprehensive AI productive imagination-understanding engine.

In our final section, we propose a training architecture that could lead to a complete AI productive imagination-understanding engine. This involves adapting the V-JEPA architecture to process out-of-order video clips and align its encoder outputs with those from encoders processing orderly, original video clips. This training structure will also include an AI cognition unit based on text descriptions that incorporate Kantian categories, ensuring all critical aspects of the coherent world model are learned.

The framework presented here is not a detailed implementation plan but a theoretical outline identifying essential components required to develop an AI system capable of building coherent world models. Given that the major components of this system have already been prototyped and tested, we anticipate that a complete technical implementation will be feasible in the future, which would mark a significant milestone in the progress of artificial general intelligence.